\documentclass[journal]{journal}

\usepackage{amsmath,amsfonts}
\usepackage{algorithmic}
\usepackage{algorithm}
\usepackage{array}
\usepackage{cite}
\usepackage{color,array,amsthm}
\usepackage{graphicx}

\ifCLASSINFOpdf
\else
\fi

\begin{document}
%
\title{
Robust Distributed Cooperative Path-Following and Local Replanning for Multi-UAVs Under Differentiated Low-Altitude Paths}
%
%
%

\author{Zimao~Sheng,
	~Zirui~Yu
	and ~Hong'an~Yang\textsuperscript{*},
\thanks{Zimao~Sheng, Hong'an~Yang is with the School of Mechanical Engineering, Northwestern Polytechnical University, Xi'an, Shaanxi, China, e-mail: (see hpShengZimao@163.com). Zirui~Yu is with Tianjin University of Science and Technology, Tianjin, China.}
\thanks{\textsuperscript{*}Corresponding author: Hong'an~Yang (e-mail: yhongan@nwpu.edu.cn).}}

\markboth{Journal of \LaTeX\ Class Files,~Vol.~6, No.~1, January~2007}%
{Shell \MakeLowercase{\textit{et al.}}: Bare Demo of IEEEtran.cls for Journals}

\maketitle
\thispagestyle{empty}

\begin{abstract}
Multiple fixed-wing unmanned aerial vehicles (multi-UAVs) encounter significant challenges in cooperative path following over complex Digital Elevation Model (DEM) low-altitude airspace, including wind field disturbances, sudden obstacles, and requirements of distributed temporal synchronization during differentiated path tracking. Existing methods lack efficient distributed coordination mechanisms for time-consistent tracking of 3D differentiated paths, fail to quantify robustness against disturbances, and lack effective online obstacle avoidance replanning capabilities. To address these gaps, a cooperative control strategy is proposed: first, the distributed cooperative path-following problem is quantified via time indices, and consistency is ensured through a distributed communication protocol; second, a longitudinal-lateral look-ahead angle adjustment method coupled with a robust guidance law is developed to achieve finite-time stabilization of path following error to zero under wind disturbances; third, an efficient local path replanning method with minimal time cost is designed for real-time online obstacle avoidance.Experimental validations demonstrate the effectiveness and superiority of the $\ $proposed strategy.
\end{abstract}


%
\IEEEpeerreviewmaketitle

\section{INTRODUCTION}

\subsection{Motivation}
\IEEEPARstart{I}{n} recent years, the multiple unmanned aerial vehicles (multi-UAVs) system has garnered widespread attention owing to its distinct advantages, including low cost, strong survivability, and excellent scalability. It plays a pivotal role in critical domains such as joint target tracking\cite{11036585}, suppression of enemy air defense\cite{10977822,SHENG2025105164}, and cooperative low-altitude rescue\cite{XING2022102972}, which aligns with the operational requirements of large-scale, high-intensity, and fast-paced multi-domain joint missions. In these practical applications, mission areas are typically characterized by complex environments modeled via Digital Elevation Models (DEM) and dotted with ground threats, encompassing undulating mountainous terrain, scattered hills, dense building clusters, and air defense fire zones.

Such environments not only pose direct risks to the flight safety of individual UAVs but also impose stringent requirements on multi-UAV cooperative operations. The core objective of cooperative path following for multi-fixed-wing UAV systems is to ensure that all UAVs accurately track predefined waypoints in complex DEM environments while achieving optimal flight speed allocation, with full consideration of each UAV’s inherent flight performance constraints. The ultimate goal is to maintain stable temporal and spatial synchronization to facilitate the efficient completion of collective tasks.

Specifically, temporal synchronization requires that multiple UAVs arrive at their respective target points as synchronously as possible. Traditional methods predominantly rely on centralized speed allocation schemes for multi-UAV systems, which impose substantial computational and communication loads on the central node (i.e., the ground base station). Furthermore, damage to the central node will result in the breakdown of the entire cooperative path following strategy of the multi-UAV system.

On the other hand, spatial safety requires that multiple UAVs minimize the risk of collisions with both the surrounding environment and other UAVs during cooperative path following. However, the dynamic variability of low-altitude environments further increases the difficulty of cooperative path following in real-world mission scenarios. For one thing, complex DEM environments may harbor unexpected temporary obstacles such as emergent low-altitude aircraft and temporarily erected facilities, which necessitates the UAV system to be equipped with real-time obstacle perception and path replanning capabilities for timely avoidance. For another, airflow disturbances in low-altitude environments, such as gusts and turbulence, directly degrade UAV flight attitudes and trajectory tracking accuracy. Collectively, these factors impose higher demands on the online robust path following capability of UAVs.

These challenges motivate us to design a novel cooperative path following control strategy. The proposed strategy is expected to enable multi-UAV systems to achieve temporal coordination and efficient obstacle avoidance via peer-to-peer communication with adjacent UAVs when tracking differentiated paths over complex DEM terrain, even in the presence of sudden obstacle threats and individual UAV malfunctions. Meanwhile, it should ensure that the cooperative following strategy exhibits favorable robustness against wind field disturbances.

\subsection{Related Work}
Current mainstream distributed cooperative path following methods for multi-UAV systems primarily focus on strategies for tracking identical or similar paths \cite{chen2021coordinated, zhang2025cooperative}. In essence, these approaches achieve multi-UAV formation control through consensus algorithms, while ensuring that each UAV’s tracking error relative to its predefined trajectory converges to zero \cite{zhao2024cooperative}. However, in DEM environments, the feasible low-altitude paths for individual UAVs vary significantly, and there remains a significant gap in research on guidance laws that enable the coordinated tracking of such differentiated paths with approximately synchronized mission completion times in the time domain.

The core challenges mainly lie in two aspects. First, it is challenging to design reference velocity profiles that can achieve temporal synchronization among UAVs under distributed cooperative communication protocols. Second, low-altitude terrain usually gives rise to additional turbulent disturbances, which impair the path following accuracy of UAVs. While existing works \cite{chen2022coordinated, ma2021path} have proposed cooperative formation following strategies for 2D curves at fixed altitudes, these methods are based solely on 2D particle models. Although the associated algorithms integrate distributed communication mechanisms and exhibit a certain degree of robustness, they cannot guarantee both the precise tracking of differentiated 3D paths in complex DEM terrain and temporal synchronization across the UAV fleet.

On another front, existing path following strategies for fixed-wing UAVs under wind disturbances, including optimal control \cite{yang2020optimal}, quasi-linear parameter-varying model predictive control (MPC) \cite{rezk2024predictive}, and look-ahead pursuit (LP) guidance laws \cite{wang2021path, kumar2024three}, can achieve robust stabilization of fixed trajectory tracking errors. Nevertheless, these methods do not quantify the robustness metrics of perturbed nonlinear error systems, let alone further optimize these metrics. Furthermore, in low-altitude path following scenarios, sudden obstacle threats are prevalent, and research on online waypoint replanning methods that can ensure smooth real-time path re-tracking while satisfying all flight constraints of UAVs remains relatively scarce.

\subsection{Contributions}
First, we formulate the distributed cooperative path-following problem of low-altitude differentiated paths under wind field disturbances from the perspective of temporal indices, and ensure temporal index consistency via a distributed communication protocol. Second, a longitudinal-lateral preview angle adjustment method is proposed to counteract wind field disturbances, which, combined with a robust guidance law, achieves finite-time convergence of path following errors to zero. Additionally, an efficient local path replanning method with minimized time cost is designed to meet the requirements of low-altitude online obstacle avoidance, with comprehensive experiments validating the effectiveness of the proposed methods.

\section{PRELIMINARIES AND PROBLEM FORMULATION}
\label{sec:PRELIMINARIES AND PROBLEM FORMULATION}
\subsection{System Model}
Consider a multi-UAV system comprising $N$ small and miniature fixed-wing UAVs (MAVs) operating at low altitude. These UAVs are tasked with tracking distinct low-altitude paths defined on a DEM, while achieving simultaneous arrival at a common target point $\textit{p}_{\text{target}}=[p_{n,\text{target}},y_{e,\text{target}},h_{\text{target}}]^T\in \mathbb{R}^3$ via local communication. Incorporating wind field disturbances, the kinematic model of $i$-th UAV is formulated as\cite{beard2012small}
\begin{equation}
	\begin{cases}
		\dot{p}_{n,i} = V_{g,i} \cos \gamma_i \cos \chi_i \\
		\dot{p}_{e,i} = V_{g,i} \cos \gamma_i \sin \chi_i  \\
		\dot{h}_i = V_{g,i}\sin\gamma_{i}\\
		\dot{\chi}_i = \frac{g}{V_{g,i}} \tan \phi_i \cos(\chi_i -\psi_i) + d_{\chi,i}\\
		\dot{\gamma}_i = \frac{g}{V_{g,i}}\left(n_{lf,i}\cos\phi_i - \cos\gamma_i \right) + d_{\gamma,i}
	\end{cases}
\end{equation}
where $p_i = [{p}_{n,i},{p}_{e,i}, {h}_i]^T$ is the position along the North, East and Down-axis in inertia coordinate frame, respectively. $ \psi_i, \gamma_{i},\chi_i, V_{g,i}$ are the yaw angle, flight path angle, course angle, ground velocity vector, respectively. $g$ is gravitational constant. $\phi_i,n_{lf,i}$ are the roll angle, load factor, respectively. Beyond wind field disturbances, the signals $V_{g,i}$, $\phi_{i}$, and $n_{lf,i}$ of $i$-th UAV must be limited as follows to ensure stable and robust flight and avoid aerial instability.
\begin{align}
	\begin{cases}
		\underline{V}_{g,i} \leq V_{g,i} \leq \overline{V}_{g,i} \\
		\underline{\phi}_{i} \leq \phi_{i} \leq \overline{\phi}_{i} \\
		\underline{n}_{lf,i} \leq n_{lf,i} \leq \overline{n}_{lf,i}	
	\end{cases}
\end{align}
 The non-predictable windy force disturbances generated by the wind field along the $\chi$ and $\gamma$-axes is $d_i = [d_{\chi,i}, d_{\gamma,i}]^T$\cite{sheng2025robustlongitudinallaterallookaheadpursuit}. 

\subsection{Communication Topology}
To enable coordination among UAVs, coordination information must be exchanged via the communication network. Within the proposed framework, each UAV exchanges coordination information exclusively with its neighbors, where the symbol $\mathcal{N}_i$ denotes the neighborhood set of the $i$-th UAV. Such interactions between neighboring UAVs are bidirectional.  
The set of adjacent nodes \(\mathcal{N}_i\) capable of communicating for $i$-th \(\text{UAV}\) determines the communication topology of the multi-UAV system. Such communication topology is often restricted by various performance constraints. Due to the payload limitations of communication modules, the communication topology of UAVs performing missions can only adopt simple structures, such as chain topology or star topology. For industrial-grade UAV data transmission radios (e.g., 433 MHz), the line-of-sight (LOS) communication range is approximately 5–20 km, while for data transmission modules (e.g., 2.4 GHz), this range is about 1–5 km. Beyond these distances, the topology will be disrupted.
\begin{figure}[htbp]
	\centering
	\includegraphics[width=0.40\textwidth]{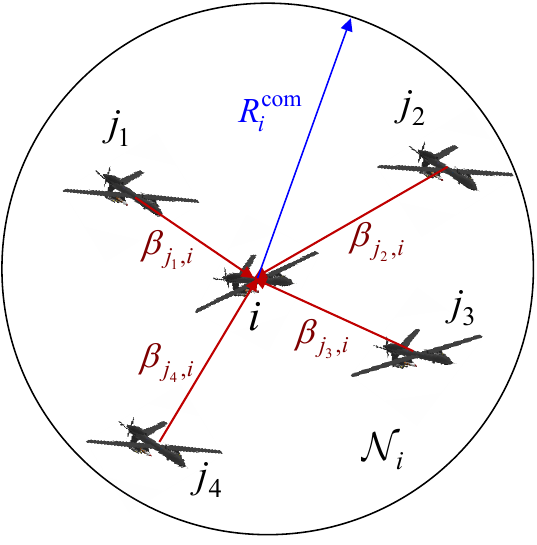}
	\caption{Communication topology between UAVs.}
	\label{fig:2}
\end{figure}
Assume that the communication radius of UAV \( i \) when attempting to communicate with other UAVs \( j \) is \( R_i^{\text{com}} \). The communication strength \( \beta_{ji} \) of each UAV \( i \) attempting to communicate with other UAVs is calculated as follows:  
\begin{align}
	\beta_{ji} = \frac{\gamma}{d_{ji}}
\end{align}
where $d_{ji}$ is the relative distance between the $i$-th UAV and $j$-th UAV, \( \gamma \) is a parameter describing the signal strength per unit distance. The set of nodes within the adjacent communication range is defined as \( \mathcal{N}_i \). The set of indices of UAVs that are within the communication range \( R_i^{\text{com}} \), sorted in descending order of signal strength, and with the number not exceeding \( C_i \) is:
\begin{align}
	\label{eq:N_i}
	\mathcal{N}_i = \left\{
	j: d_{ji} \leq R_i^{\text{com}}, \beta_{k,i}\geq \beta_{k+1,i}, k\in \mathcal{N}_i, |\mathcal{N}_i|\leq C_i 
	\right\}
\end{align}
These constraints impose two key limitations on UAVs: they can only communicate with adjacent UAVs within a radius of \(R_i^{\text{com}}\), and the number of other UAVs they can communicate with is limited to a maximum number. For scenarios where the number of adjacent nodes exceeds this maximum, the most suitable nodes in \(C_i\) should be selected based on the principle of "nearest adjacent-node priority for communication" to avoid congestion. 

\subsection{Problem Formulation }
This section focuses on developing a robust path-following solution for fixed-wing UAVs in the presence of disturbances such as wind fields and model simplifications. For the \(i\)-th UAV, it is assumed to adhere to a set of generic, predefined reference waypoints denoted as \( \mathcal{P}_i=\left\{ p_{n,i,c}^{(j)}, p_{e,i,c}^{(j)}, h_{i,c}^{(j)}\right\}_{j=1}^n \). The look-ahead pursuit path-following control for UAVs formulates the predefined path-following problem of \( \mathcal{P}_i \) as a tracking problem for moving virtual waypoints. The virtual target point 
\begin{align}
	y_{c,i} = \left[ p_{n,i,c}^{*}, p_{e,i,c}^{*}, h_{i,c}^{*}\right]^T 
\end{align}
traverses along the reference path \( \mathcal{P}_i \), with a Line-of-Sight (LOS) vector extends from the UAV to this point. Specifically, the virtual target point \( y_{c,i} \) is defined as the nearest point located ahead along the predefined $\mathcal{P}_i$ in the UAV’s velocity direction\cite{wang2021path}.
\newtheorem{assumption3}{ASSUMPTION}
\begin{assumption3}
	It is assumed that at any arbitrary instant, the \(i\)-th UAV can obtain measurements of position and attitude by means of sensors including GPS, barometers, airspeed sensors, and inertial measurement units (IMUs), and derive the optimal estimate \( \hat{x}_{i} \) of the current state \( x_{i} \) through the Extended Kalman Filter (EKF) for observation $y_i$. Additionally, this estimate is assumed to be unbiased, i.e. ${x}_i \approx \hat{x}_i$.
\end{assumption3}

\newtheorem{problem1}{PROBLEM}
\begin{problem1}
	\label{pr:problem2}
	(Distributed cooperative path-following)\\ The distributed cooperative path-following of multi-UAVs is said to achieve attaining simultaneous arrival at the target \( p_{\text{target}} \), i.e., $\lim_{t\rightarrow\infty}\left|
	\theta_i -\theta_j  
	\right|=0$ for any UAVs $i,j$, through a distributed communication protocol for each reference \( V_{g,c,i}(t) \) is regulated, and
	\begin{align}
		\label{eq:theta}
		\theta_i = T_i(p_{\text{target}}) = \frac{
			d(p_i,y_{c,i}) + l_{\mathcal{P}}(y_{c,i},p_{\text{target}}) 
		}{V_{g,i}}
	\end{align}
	Here, \( T_i(p_{\text{target}}) \) can be approximately estimated as the time from location of $i$-th UAV to $p_{\text{target}}$. \( l_{\mathcal{P}}(y_{c,i}, p_{\text{target}}) \) denotes the  length from the starting waypoint \( y_{c,i} \) to the terminal waypoint \( p_{\text{target}} \) at the trajectory \( \mathcal{P} \). 
\end{problem1}
\begin{figure}[htbp]
	\centering
	\includegraphics[width=0.48\textwidth]{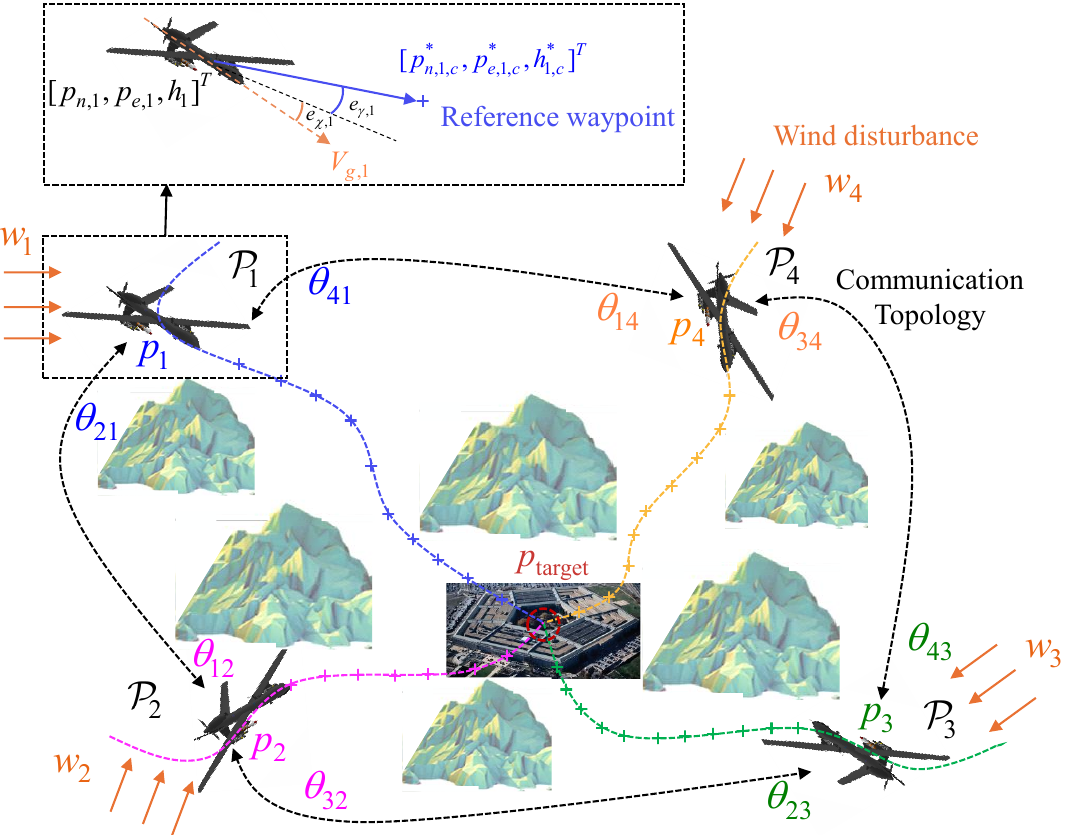}
	\caption{Schematic diagram of distributed cooperative path following for multi-UAVs system.}
	\label{fig:1}
\end{figure}
The objective of distributed cooperative path-following problem is to design an appropriate coordination reference groundspeed command $V_{g,c,i}$ for $i$-th UAV, which can be effectively solved in a distributed optimization manner.

\section{DISTRIBUTED COOPERATIVE ROBUST PATH-FOLLOWING AND LOCAL REPLANNING}

\subsection{Robust Stabilization for Path-Following Error}
During the process of the UAV tracking along the current waypoint $y_{c,i}$, the position errors along the North, East, Height-axis directions are defined as
\begin{align}
	\footnotesize
	e_{p_n,i} = p_{n,i,c}^* - p_{n,i},
	e_{p_e,i} = p_{e,i,c}^* - p_{e,i},
	e_{h,i} = h_{i,c}^* - h_{i}
\end{align}
We propose the following assumption about the general structure of the UAV flight control system:
\newtheorem{assumption}[assumption3]{ASSUMPTION}
\begin{assumption}
	Each UAV $i$'s flight control system (autopilot) is capable of robustly tracking the reference load factor signal $n_{lf,c,i}$, roll angle signal $\phi_{c,i}$, and reference groundspeed $V_{g,c,i}$ \cite{rezk2024predictive}. 
	
\end{assumption}
As the core of this study, an robust path-following
controller is designed in this section. The objective of this paper is to design the desired control commands $\phi_{c,i}$, $n_{lf,c,i}$ and $V_{g,c,i}$ to ensure that the path following errors can be robust stablized even when subject to bounded path angles disturbances. The sufficient conditions for
ensuring the fast finite-time stability of longitudinal-lateral
look-ahead pursuit guidance law are derived. With the lateral flight path angle $\chi_{c,i}$, longitudinal flight path angle $\gamma_{c,i}$ of the \(i\)-th UAV given by 
\begin{align}
	&\chi_{c,i} = \tan^{-1}\frac{p_{e,i,c}^{*} - p_{e,i}}{p_{n,i,c}^{*} - p_{n,i}}\\ &\gamma_{c,i} = \tan^{-1}\frac{h_{i,c}^{*} - h_{i}}{\sqrt{(p_{e,i,c}^{*} - p_{e,i})^2 + (p_{n,i,c}^{*} - p_{n,i})^2}}	
\end{align}
and 
$V_{g,i}$ is determined by the distributed temporal coordination protocol. During the process of tracking a virtual target point,
its lateral look-ahead angle $\eta^{lat}_i$ and longitudinal look-ahead angle $\eta^{lon}_i$ can be defined as
\begin{align}
	\eta^{lat}_i = \chi_{c,i} - \chi_i,\quad \eta^{lon}_i = \gamma_{c,i} - \gamma_{i}
\end{align}
where $\eta^{lat}_i$ and $\eta^{lon}_i$ are limited within the certain ranges as follows:
\begin{align}
	\underline{\eta}^{lat} \leq \eta^{lat}_i \leq \overline{\eta}^{lat},\quad \underline{\eta}^{lon} \leq \eta^{lon}_i \leq \overline{\eta}^{lon}
\end{align}
In our previous work Ref.\cite{sheng2025robustlongitudinallaterallookaheadpursuit}, we demonstrated that the look-ahead angle pairs $(\eta^{lat}_i, \eta^{lon}_i)$ do not necessarily need to be stabilized to zero. Even when $\eta^{lat}_i$ and $\eta^{lon}_i$ are within a disturbance range, the path following errors $e_{p_n,i}, e_{p_e,i}, e_{h,i}$ can be stabilized to zero in finite time, which indicates that the system has good robustness.
\newtheorem{theorem1}{THEOREM}
\begin{theorem1}
	If for any reference waypoint $y_{c,i}\in\mathcal{P}_i$ is constrained by $\sqrt{\dot{p}_{n,i,c}^{*2} + \dot{p}_{e,i,c}^{*2}+ \dot{h}_{i,c}^{*2} }\leq \mathcal{L}_c$, $\gamma_i\left( h_i - h_{i,c}^{*} \right)\leq 0$, and there exists $ 0\leq \delta^{lon},\delta^{lat} < \frac{\pi}{2}$, here $V_g\cos\delta^{lon}\cos\delta^{lat}>\mathcal{L}_c $
	 such that $|\eta^{lon}_i|\leq \delta^{lon}, 
	 |\eta^{lat}_i|\leq \delta^{lat}$. Then $e_{p_n,i}, e_{p_e,i}, e_{h,i}$ will be stablized to 0 within a finite time $T(e_0)$, where $e_0$ is the initial position error\cite{sheng2025robustlongitudinallaterallookaheadpursuit}.
\end{theorem1}
Further, the specific form of the adopted reference roll angle $\phi_{c,i}$ and load factor $n_{lf,c,i}$ is expressed by:
\begin{align}
	\label{eq:phi_n_lf}
	\begin{split}
		&\phi_{c,i} = -\sin^{-1}\left(
		\frac{V_{g,i}\cos\phi_i }{g}f_{\chi}\left(\eta^{lat}_i,\eta^{lon}_i\right)
		\right) \in \left[\underline{\phi}_i, \overline{\phi}_i\right]\\
		&n_{lf,c,i} = \frac{g\cos\gamma_i - V_{g,i} f_{\gamma}\left( \eta^{lat}_i,\eta^{lon}_i\right) }{g\cos\phi_{c,i}} \in \left[\underline{n}_{lf,i}, \overline{n}_{lf,i}\right]
	\end{split}
	\end{align}
where the $\eta_i = [\eta^{lat}_i,\eta^{lon}_i]^T$, and $f(\eta_i) = [f_{\chi}(\eta_i),f_{\gamma}(\eta_i)]^T$ determine the specific formulation of the guidance law,
\begin{align}
	\lambda_{\max}\left(
	\left[\frac{\partial f(0)}{\partial \eta_i} \right]^T + 
	\left[ \frac{\partial f(0)}{\partial \eta_i} \right]
	\right) < 0
\end{align}
here we set $f_{\chi}(\eta_i) = -k_{\chi}\sin\eta^{lat}_i$, $f_{\gamma}(\eta_i) = -k_{\gamma}\sin\eta^{lon}_i$, $k_{\chi},k_{\gamma}>0$ are the gain coefficients.

\subsection{Efficient Local Path Replanning}
\begin{figure}[htbp]
	\centering
	\includegraphics[width=0.48\textwidth]{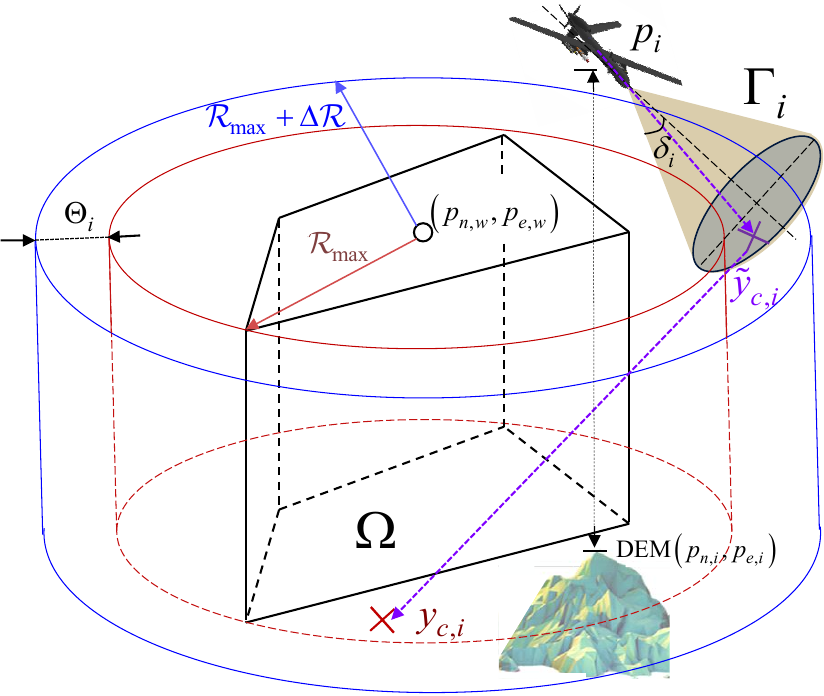}
	\caption{Schematic diagram of the $i$-th UAV sampling a new waypoint \(\tilde{y}_{c,i}\) within the local region \(\Gamma_i\)}
	\label{fig:3}
\end{figure}

As shown in Fig.\ref{fig:3}, when the \(i\)-th UAV is tracking the virtual waypoint \(y_{c,i}\), if a sudden obstacle threat \(\Omega\) blocks its path, it is necessary to perform multiple random samplings ahead of the original waypoint \(y_{c,i}\) to locally replan a new optimal waypoint \(\tilde{y}_{c,i}\) as the current target waypoint. Repeating this process multiple times allows us to sample a series of optimal waypoints, which serve as the UAV's current local replanned path.

\newtheorem{assumption4}[assumption3]{ASSUMPTION}
\begin{assumption4}
	Assuming each UAV can detect the contour coordinates of the current obstacle \(\Omega\) via its detection radar, it can further deduce the lateral coordinates of the obstacle’s center of gravity as \((p_{n,w}, p_{e,w} )\) and measure the altitude $\text{DEM}\left(p_{n,i}, p_{e,i} \right)$ under the current terrain using its sensors. Thus, based on the center of gravity coordinates of \(\Omega\), the maximum lateral expansion radius of the region can be calculated by
	\begin{align}
		\bar{R} = \sup_{\left(p_n,p_e,h\right)\in \Omega}
		d_{lat}\left(
		\begin{bmatrix}
			p_n \\ p_e
		\end{bmatrix},
		\begin{bmatrix}
			p_{n,w} \\ p_{e,w}
		\end{bmatrix}
		\right)
	\end{align}
	where \(d_{lat}(a,b)\) denotes the lateral distance between coordinates \(a\) and \(b\).  
\end{assumption4}
Given the search radius increment \(\Delta R\), included angle \(\delta_i\) and the \(i\)-th UAV’s altitude increment \(\Delta h_i\), the optimal new waypoint \(\tilde{y}_{c,i}\) is sought in the the following feasible regions $\Gamma_i$:  
\begin{align}
	\label{eq:Gamma_i}
	\Gamma_i = \left\{
	(p_n,p_e,h)\in \Theta_i:
	\left<\mu_i
	,
	\begin{bmatrix}
		p_n - p_{n,i} \\
		p_e - p_{e,i} \\
		h- h_{i} 
	\end{bmatrix}
	\right>\leq \delta_i
	\right\}
\end{align}
where \(\left<a,b \right>\) denotes the included angle between vector $a$ and $b$, $\mu_i= \left[ \cos \gamma_i\cos \chi_i, \cos \gamma_i\sin \chi_i, \sin \gamma_i \right]^T$ is the velocity vector of the UAV,
\begin{align}
	\begin{split}
		\Theta_i = \big \{& (p_n,p_e,h):
		\bar{R} \leq d_{lat}\left(
		\begin{bmatrix}
			p_n \\ p_e
		\end{bmatrix},
		\begin{bmatrix}
			p_{n,w} \\ p_{e,w}
		\end{bmatrix}\right) \leq \bar{R}  + \Delta R\\
		&\text{DEM} \left(p_{n,i},p_{e,i} \right) \leq h \leq
		\text{DEM} \left(p_{n,i},p_{e,i} \right) + \Delta h_i
		\big \}
	\end{split}
\end{align}
According to Ref.\cite{sheng2025robustlongitudinallaterallookaheadpursuit}, the upper bound of the flight time from the UAV's coordinate \(p_i\) to \(\tilde{y}_{c,i}\) (denoted as \(T(p_i, \tilde{y}_{c,i})\)), and the upper bound of the flight time from \(\tilde{y}_{c,i}\) to \(y_{c,i}\) (denoted as \(T(\tilde{y}_{c,i}, y_{c,i})\)) can be described as:
\begin{align}
	\begin{split}
		&T(p_i, \tilde{y}_{c,i}) = \frac{d_*\left( 
			p_i, \tilde{y}_{c,i}
			\right)}{V_{g,i}\cos \eta_{1}^{lon}\cos \eta_{1}^{lat} }\\
		&T(\tilde{y}_{c,i}, y_{c,i}) = \frac{d_*\left( 
			\tilde{y}_{c,i}, y_{c,i}
			\right)}{V_{g,i}\cos \eta_{2}^{lon}\cos \eta_{2}^{lat} }
	\end{split}
\end{align}
where \(d_*(a,b)\) denotes the 3D distance between coordinates \(a\) and \(b\), $\eta_1^{lat},\eta_1^{lon}$ can be described as
\begin{align}
	\begin{split}
		&\eta_1^{lat} = \tan^{-1}\frac{\tilde{p}_{e,i} - {p}_{e,i}}{\tilde{p}_{n,i} - {p}_{n,i}} - \chi_i \\
		&\eta_1^{lon} = \tan^{-1}\frac{ \tilde{h}_{i} - {h}_{i} }{\sqrt{\left(\tilde{p}_{e,i} - {p}_{e,i} \right)^2 + \left(\tilde{p}_{n,i} - {p}_{n,i} \right)^2} } - \gamma_i
	\end{split}
\end{align}
and $\eta_2^{lat},\eta_2^{lon}$ can be approximately estimated as
\begin{align}
	\begin{split}
		\eta_2^{lat} \approx  &\tan^{-1}\frac{{p}_{e,i}^* - \tilde{p}_{e,i}}{{p}_{n,i}^* - \tilde{p}_{n,i}} -
		\tan^{-1}\frac{\tilde{p}_{e,i} - {p}_{e,i}}{\tilde{p}_{n,i} - {p}_{n,i}}\\
		\eta_2^{lon} \approx 
		&\tan^{-1}\frac{ {h}^*_{i} - \tilde{h}_{i} }{\sqrt{\left({p}_{e,i}^* - \tilde{p}_{e,i} \right)^2 + \left( {p}_{n,i}^* - \tilde{p}_{n,i} \right)^2} }\\
		&- \tan^{-1}\frac{ \tilde{h}_{i} - {h}_{i} }{\sqrt{\left(\tilde{p}_{e,i} - {p}_{e,i} \right)^2 + \left(\tilde{p}_{n,i} - {p}_{n,i} \right)^2} }
	\end{split}
\end{align}
We aim to find \(\tilde{y}_{c,i}^*\) that minimizes the total time \(T(p_i, \tilde{y}_{c,i}) + T(\tilde{y}_{c,i}, y_{c,i})\) through \(K\) random explorations within the feasible region \(\Gamma_i\), i.e., 
\begin{align}
	\footnotesize
	\label{eq:y_c_i_star}
	\tilde{y}_{c,i}^* = \text{argmin}_{\tilde{y}_{c,i}\in \Gamma_i}
	\left\{
	\frac{d_*\left( 
		p_i, \tilde{y}_{c,i}
		\right)}{\cos \eta_{1}^{lon}\cos \eta_{1}^{lat} }
	+
	\frac{d_*\left( 
		\tilde{y}_{c,i}, y_{c,i}
		\right)}{\cos \eta_{2}^{lon}\cos \eta_{2}^{lat} }
	\right\}
\end{align}
This process is repeated iteratively until the obstacle-free waypoints sequence \(Y_c\) relative to the target point \(y_{c,i}\) is obtained. The specific code is provided in Alg. \ref{alg:local_path_replanning}.
\begin{algorithm}[htbp]
	\caption{Pseudocode for local path replanning}
	\label{alg:local_path_replanning}
	\begin{algorithmic}[1]
		\REQUIRE $i$-th UAV's coordinate $p_i$, target point coordinate $y_{c,i}$, obstacle $\Omega$, parameters $K, \Delta R, \Delta h_i,\delta_i$.
		\ENSURE The waypoints sequence \(Y_c\).  
		\STATE Initialize $Y_c = \emptyset$.
		\WHILE{The straight-line between \(p_i\) and \(y_{c,i}\) is obstructed by \(\Omega\)}
		\STATE Safe feasible region \(\Gamma_i\) is determined by (\ref{eq:Gamma_i}).
		\STATE Calculate the optimal waypoint \(\tilde{y}_{c,i}^*\) via (\ref{eq:y_c_i_star}) through \(K\) random explorations.
		\STATE $Y_c\leftarrow Y_c \cup \tilde{y}_{c,i}^*$.
		\STATE $p_i \leftarrow \tilde{y}_{c,i}^*$.
		\ENDWHILE
	\end{algorithmic}
\end{algorithm}
\begin{algorithm}[htbp]
	\caption{Pseudocode for distributed cooperative robust path-following controller}
	\label{alg:robust path-following}
	\begin{algorithmic}[1]
		\REQUIRE The virtual waypoint \( y_{c,i} \) to be path-followed.
		\ENSURE Reference inputs $u_i$.  
		\FOR{$\text{UAV}\ i=1,2\ldots N$}
		\STATE The $i$-th UAV determines the current virtual waypoint \( y_{c,i} \) to be path-followed.
		\IF{The path from the UAV to \( y_{c,i} \) is blocked by the obstacle \(\Omega\)}
		\STATE Call the Alg.\ref{alg:local_path_replanning} to obtain the new waypoint sequences \(Y_c\), and let $y_{c,i} \leftarrow Y_c(1)$.
		\ENDIF
		\STATE Establish the communication topology \(\mathcal{N}_i\) for the \(i\)-th UAV via  (\ref{eq:N_i}).
		\STATE Compute the cooperativity metric $\theta_i$ by (\ref{eq:theta}).
		\STATE Compute the $V_{g,c,i}$ via (\ref{eq:coordination_protocol})-(\ref{eq:V_a_i_c}).
		\STATE Compute the $\phi_{c,i}$ and $n_{lf,c,i}$ via (\ref{eq:phi_n_lf}).
		\ENDFOR
	\end{algorithmic}
\end{algorithm}

\subsection{Distributed Temporal Coordination Protocol}
To achieve the coordination objective in PROBLEM \ref{pr:problem2}, under the guarantee of ASSUMPTION \ref{ass:assumption5} regarding the communication topology, we adopt the following purely distributed coordination control law mentioned in Ref. \cite{CHEN2022260} as follows 
\begin{align}
	\label{eq:coordination_protocol}
	\dot{\theta}_{i} = -\sum_{j\in\mathcal{N}_i} \beta_{ji}\tanh\left(
	k_{\theta}\left(
	\theta_i - \theta_j
	\right)
	\right) +\gamma_d
\end{align}
where $k_{\theta}, \gamma_d$ are positive constants proposed based on the consensus theory, and $\gamma_d$ is used to constrain the the progression rate. This coordination strategy in the paper can achieve leaderless synchronization and is thus more robust to malfunctions of certain UAVs.  
\newtheorem{assumption5}[assumption3]{ASSUMPTION}
\begin{assumption5}
	\label{ass:assumption5} 
	In this paper, we assume that the communication topology connectivity satisfies the "persistency of excitation" (PE-like) condition\cite{4287131}. These assumptions can be readily ensured in practice and are widely adopted in the literature. This condition permits the communication topology to lose connectivity, provided that connectivity is established in the integral sense.  
\end{assumption5}
when adopting the distributed coordination strategy proposed in (\ref{eq:coordination_protocol}), this will ensure that \(|\theta_i - \theta_j| \to 0\) and \(\dot{\theta}_i \to \gamma_d\) for any \(i\)-th and \(j\)-th UAVs in the communication topology of the multi-UAV system\cite{CHEN2022260}. Subject to the aforementioned communication protocol, further discretization results in the ideal reference \(\theta_{i,c}\) as 
\begin{align}
	\theta_{i,c} = \theta_i + \dot{\theta}_i \Delta t	
\end{align}
where \(\Delta t\) denotes the sampling time. We further update the current reference airspeed input for the \(i\)-th UAV using \(\theta_{i,c}\) as follows:  
\begin{align}
	\label{eq:V_a_i_c}
	V_{g,c,i} = \text{clip}\left(
	V_{g,i} - k_{V_g}\left( 
	\theta_{i,c} - \theta_i	\right), \underline{V}_{g,i},
	\overline{V}_{g,i}
	\right)
\end{align}
where \( k_{V_g} > 0 \) is a coefficient for tuning the acceleration of the reference speed update, and \( \text{clip}(x, a, b) \) is a clipping function restricting \( x \) to \([a, b]\).  

In summary, the complete pseudocode of distributed cooperative robust path-following controller of multi-UAVs system is shown in Alg. \ref{alg:robust path-following}.

\section{SIMULATION}
\begin{figure}
	\includegraphics[width=0.48\textwidth]{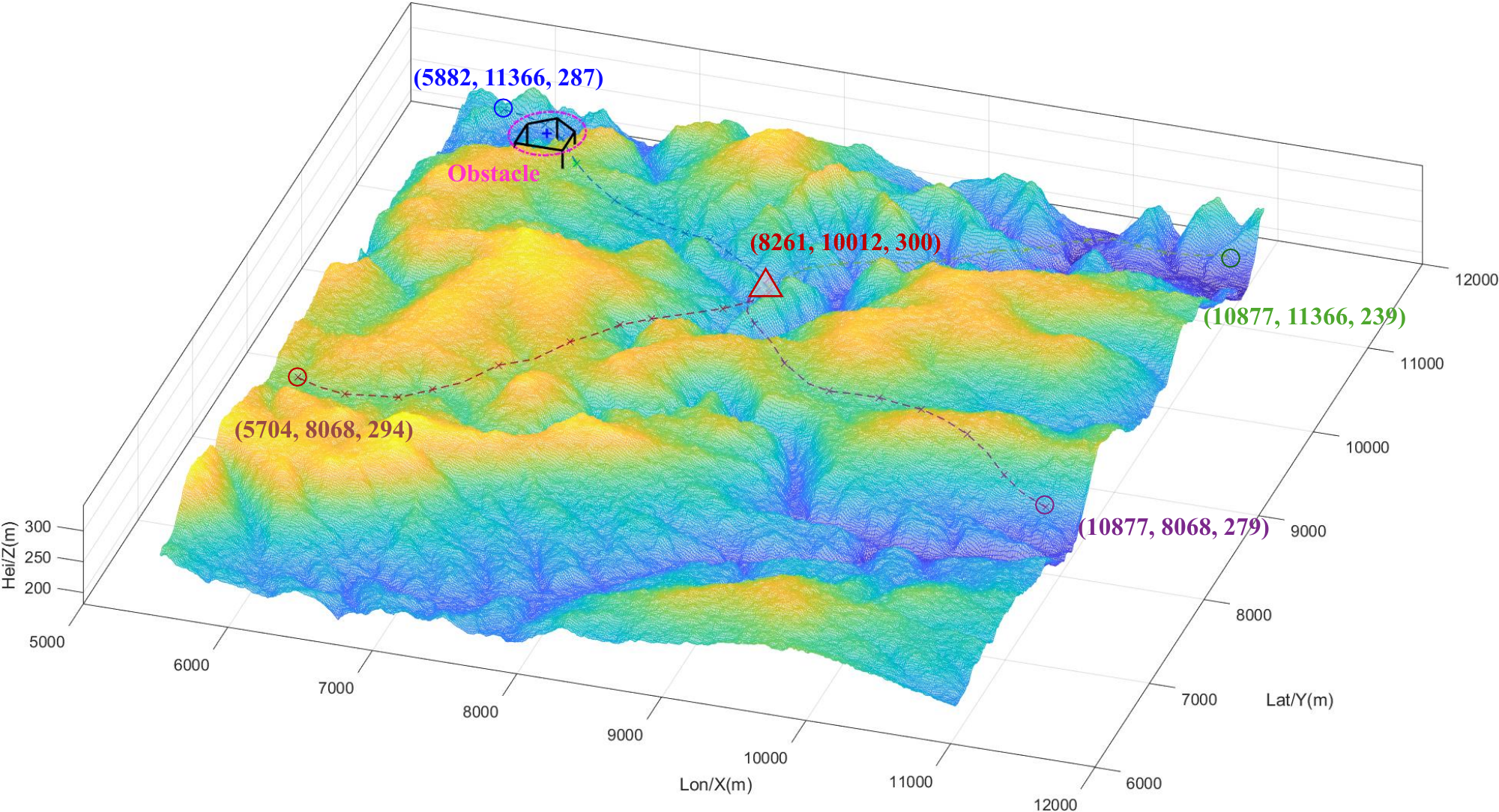}
	\caption{Starting coordinates, target coordinates $p_{target}$, and path $\mathcal{P}_i$ to be followed for each $\text{UAV}_i$.}
	\label{fig:map}
\end{figure}

In this section, we verify the effectiveness and scalability of the proposed algorithm for path following in low altitude complex terrain with obstacle. The experimental environment is configured as a low altitude path following experiment. The waypoints of the path, respective starting points and target points, designed to mimic real-world navigation tasks, are shown in Fig.\ref{fig:map}, featuring sharp turns and altitude changes. The UAV model required for simulations is a small Miniature Air Vehicle (MAV), Aerosonde, specifically a fixed wing configuration with a 1.2 m wingspan, and its physical and aerodynamic parameters specified below are given in Ref. \cite{beard2012small}. 

Meanwhile, we also apply wind field disturbances. Here, the non steady gust portion is modeled using white noise filtered through a linear time invariant system based on the von Kármán turbulence spectrum \cite{stengel2005flight} within the Dryden transfer functions \cite{beard2012small}. The Dryden gust model parameters specified in MIL-F-8785C \cite{langelaan2011wind} are: turbulence intensities along body frame axes \( \sigma_u = \sigma_v = 2.12 \, \text{m/s} \), \( \sigma_w = 1.4 \, \text{m/s} \); spatial wavelengths \( L_u = L_v = 200 \, \text{m} \), \( L_w = 50 \, \text{m} \). The ambient wind is typically expressed in the inertial frame as \( V_{w_s} = [2.5, 0, 0]^T \). The influence mechanisms of specific aerodynamic forces and moments are detailed in Ref.\cite{beard2012small}.
A cooperative path following experiment of 4 UAVs is carried out on the DEM shown in Fig.\ref{fig:map}, and a sudden obstacle as shown in the figure is deployed on the waypoints of the 1st UAV to test its online path replanning capability. We set \(C_i = 2\) for each UAV, and the upper limit of their mutual communication range \(R_i^{com} = 30\ \text{km}\). The sampling time \(\Delta t\) is set as 1 s, the flight speed range \([\underline{V}_{g,i}, \overline{V}_{g,i}]\) is 9–18 m/s, the roll angle range \([\underline{\phi}_i,\overline{\phi}_i]\) is \(-0.6\ \text{rad}\) to \(0.6\ \text{rad}\), and the load factor range \([\underline{n}_{lf,i},\overline{n}_{lf,i}]\) is 0 to 2.1. Configurations of other algorithm parameters: \(k_{\chi}=k_{\gamma}=8.8844\), \(K=2000\), \(\Delta \mathcal{R}=500\ \text{m}\), \(\Delta h_i=20\ \text{m}\), \(k_{\theta}=1\), \(k_{V_g}=0.001\).
\begin{figure}
	\includegraphics[width=0.46\textwidth]{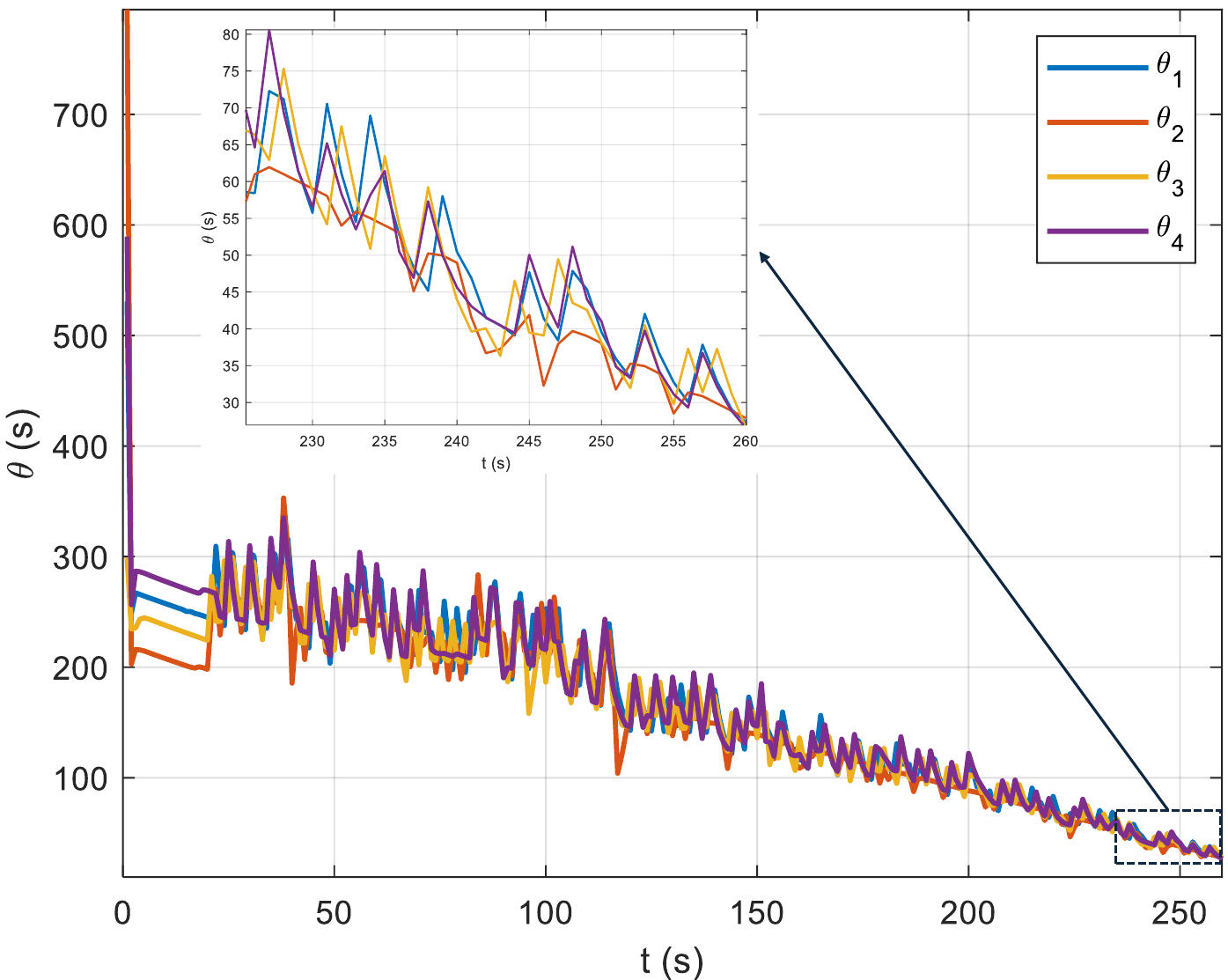}
	\caption{Time-series curves of \(\theta_i\) for each UAV $i$ during 0s - 260s.}
	\label{fig:coordinated}
\end{figure}
\subsection{Validation of Effectiveness}
\begin{figure*}
	\begin{tabular}{cc}
		\includegraphics[width=0.48\textwidth]{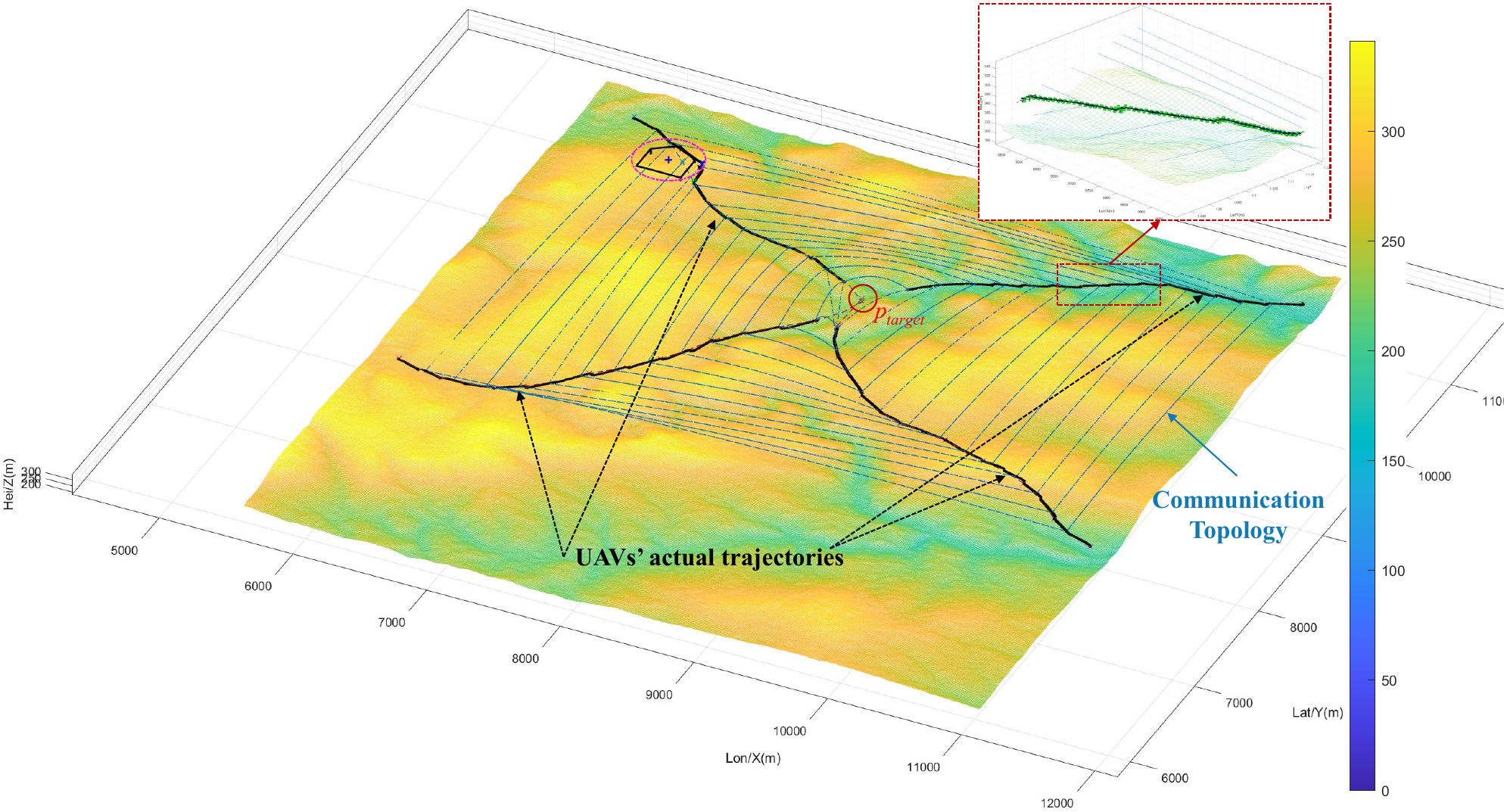} &
		\includegraphics[width=0.48\textwidth]{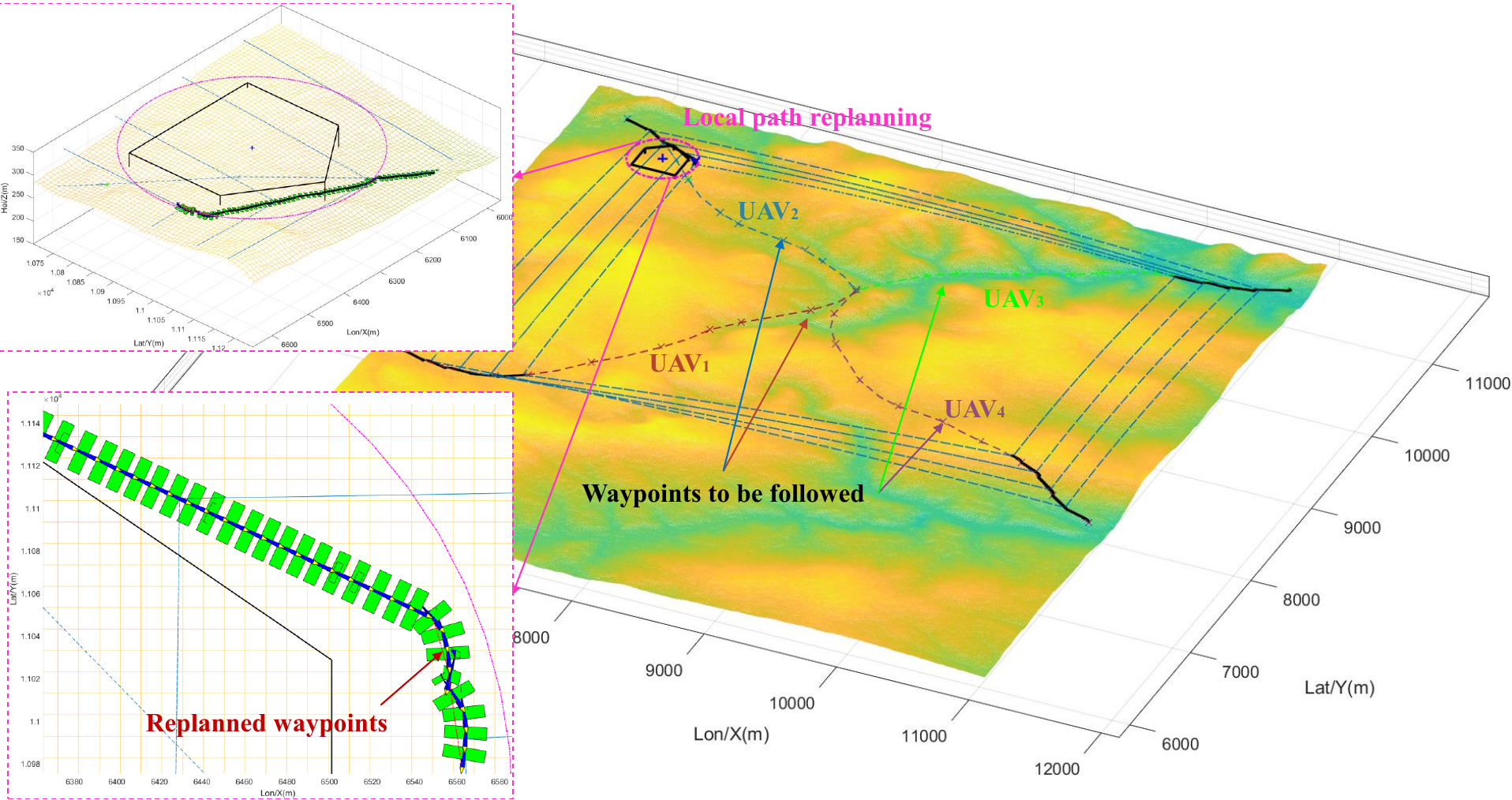}
		\\
		(a)  &
		(b) 
	\end{tabular}
	\caption{Robust cooperative path-following and local path replanning process of the multi-UAV system. (a): Simulation of the cooperative path-following process from 0s to 260s; (b): Simulation of the local path replanning at 75s.}
	\label{fig:complete_replanned}
\end{figure*}

We select the cooperative path following simulation of the multi-UAVs system from 0 s to 260 s, and the experimental results are shown in Fig. \ref{fig:complete_replanned}. Subfigure (a) shows that each UAV can achieve accurate and robust following of the low altitude path under wind field disturbances. In addition, subfigure (b) shows that at 75 s, emergency online path replanning for the sudden obstacle can be completed within no more than 2 s, and the obstacle is bypassed smoothly. This process is realized by the switched accurate path following of the replanned waypoints.

On the other hand, regarding the coordinated consistency of the time index \(\theta_i\), referring to Fig.\ref{fig:coordinated}, it can be found that even after path replanning, through the interaction of the distributed communication topology, the time indices of multiple UAVs still show a good fitting effect at the final moment, and the maximum difference of \(\theta_i\) among all UAVs does not exceed 15 s, which indicates that good consistency is achieved.
\begin{figure*}
\centering
\begin{tabular}{ccc}
	\includegraphics[width=0.29\textwidth]{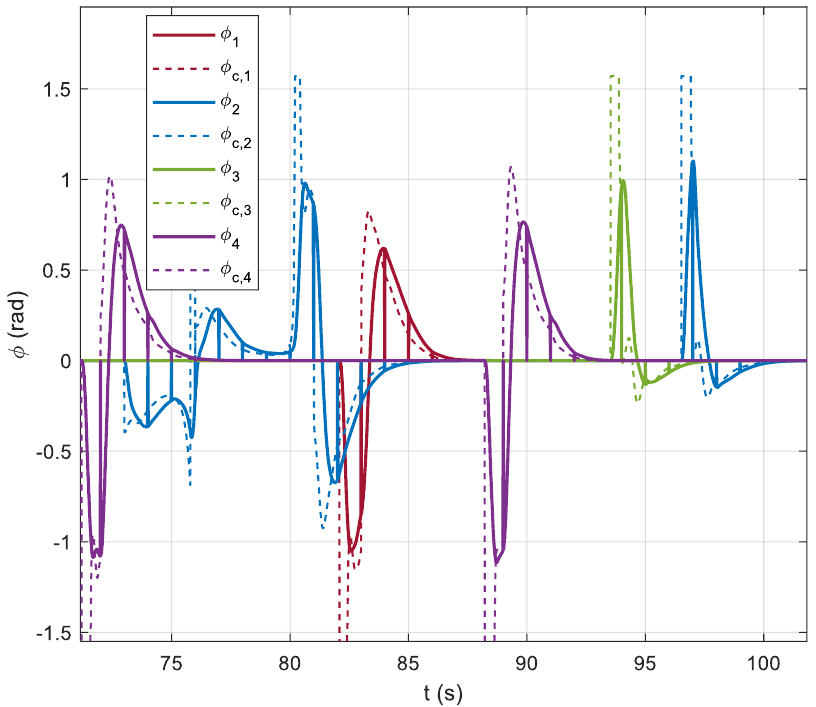}
	&
	\includegraphics[width=0.29\textwidth]{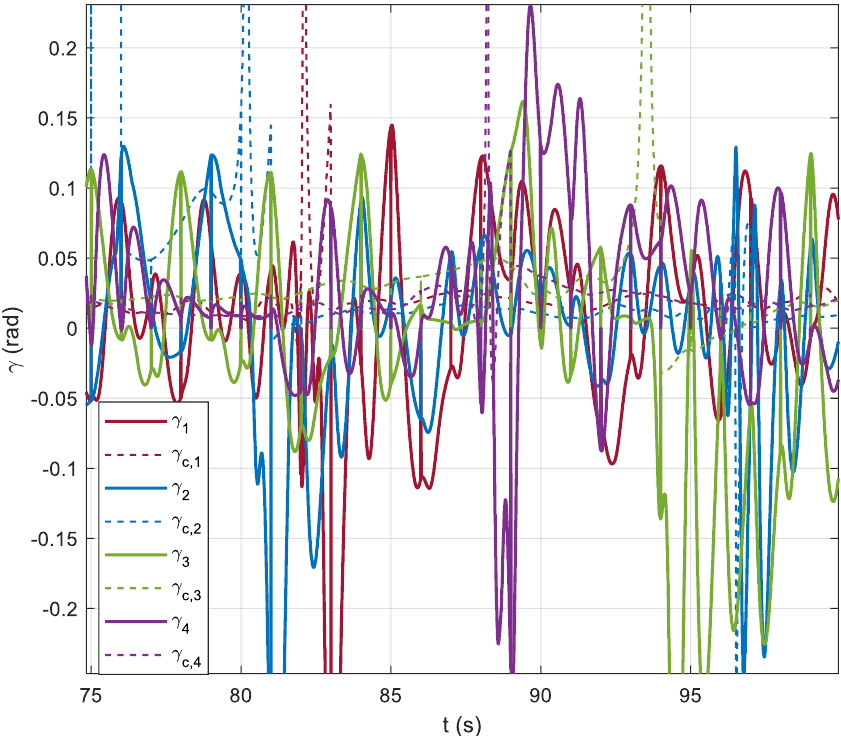}
	&
	\includegraphics[width=0.34\textwidth]{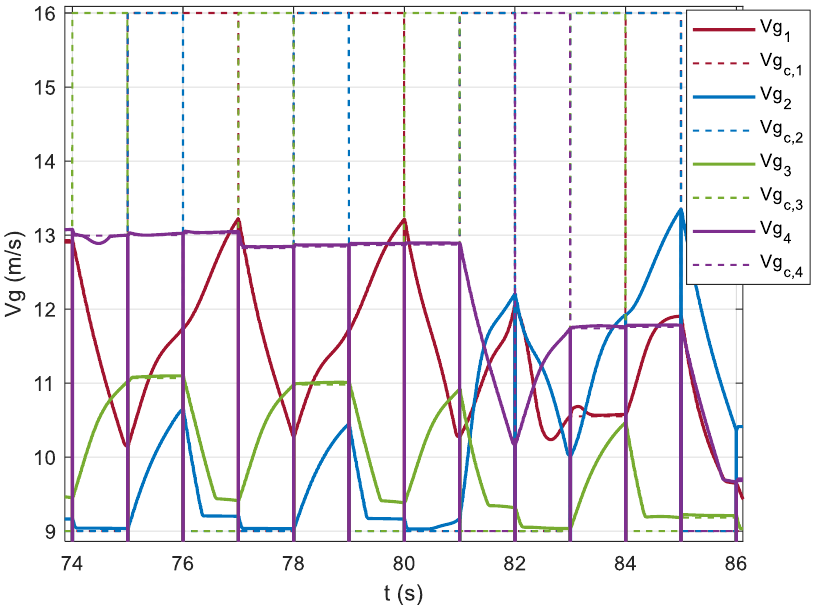}\\
	(a) & (b)  & (c) 
\end{tabular}	
\caption{Tracking performance of the autopilot for the reference commands generated by the guidance law during 75s – 100s. (a): Signal tracking curves of $\phi_{c,i}$ for each UAV $i$; (b): Signal tracking curves of $\gamma_{c,i}$ for each UAV $i$; (c): Signal tracking curves of $V_{g,c,i}$ for each UAV $i$.}
\label{fig:result2}
\end{figure*}

In the tracking process of the autopilot, we investigated the reference signal tracking effects of \(\phi_{c,i}\), \(\gamma_{c,i}\), and \(V_{g,c,i}\) for each UAV \(i\) from 75 s to 100 s, as shown in Fig.\ref{fig:result2}. Good signal tracking is achieved for \(\phi_{c,i}\), and the maximum tracking error for \(\gamma_{c,i}\) does not exceed 0.15 rad, which indicates the effectiveness of the autopilot in tracking its reference commands, verifying the feasibility of its actual deployment. 

\subsection{Validation of Scalability}

\begin{table}
	\centering
	\caption{The comparison of $\bar{\text{AE}}$, $\bar{\text{RMSE}}$, MD and replanning time for different number of UAVs from 0s to 100s.}
	\begin{tabular}{ccccc}
		\hline
		Number of UAVs & $\bar{\text{AE}}(\text{m})$ &
		$\bar{\text{RMSE}}(\text{m})$ & MD(s) & RT(s)\\
		\hline
		4 & 9.7817 & 26.7499 & 13.9428 & 3.7138\\
		7 & 4.3812 & 11.4767 & 25.5534 &3.653\\
		10 & 3.1742 & 9.1567 & 9.8762 & 2.9872\\
		13 & 4.5678 & 12.3907 & 15.5623 & 2.5940\\
		\hline
	\end{tabular}
	\label{tab:tabx}
\end{table}
To verify the scalability of the proposed algorithm in scenarios with a larger fleet of UAVs, we compared the performance metrics of the system under four different UAV quantities (i.e., 4, 7, 10, and 13 UAVs) over the time interval of 0–100 s. The selected metrics include the Average Error
\begin{align}
	\footnotesize
	\bar{\text{AE}}=\frac{1}{N}\sum_{i=1}^N\sum_{j=1}^{|\mathcal{P}_i|}\sqrt{e_{p_n,i}^2(t_j) +e_{p_e,i}^2(t_j) +e_{h,i}^2(t_j)}
\end{align}
Average Root Mean Square Error, 
\begin{align}
	\bar{\text{RMSE}} = \frac{1}{N}\sum_{i=1}^N\sqrt{\frac{1}{|\mathcal{P}_i|-1}\sum_{j=1}^{|\mathcal{P}_i|}\left( \left\| e_i(t_j)\right\| - \left\| \bar{e}_i\right\| \right)^2}
\end{align}
Herein, \(t_j\) denotes the time instant when UAV \(i\) is closest to the waypoint \(y_{c,i}\), \(e_i(t_j)= [e_{p_n,i}(t_j),e_{p_e,i}(t_j),e_{h,i}(t_j)]^T\), and \(\bar{e}_i =\frac{1}{|\mathcal{P}_i|} \sum_{j=1}^{|\mathcal{P}_i|}e_i(t_j)\), where \(|\mathcal{P}_i|\) represents the number of waypoints on the path \(\mathcal{P}_i\).
Maximum \(\theta\) Deviation (\(\text{MD} = \max_{i,j}\left\| \theta_{i} - \theta_j\right\|\)) and Replanning Time (\(\text{RT}\)). Specifically, \(\bar{\text{AE}}\) and \(\bar{\text{RMSE}}\) are employed to characterize the robustness of UAV path following \cite{10517899}; \(\text{MD}\) quantifies the degree of agreement of the consistency metric; and \(\text{RT}\) is used to evaluate the real-time performance of the path replanning process.
In each experiment, the initial coordinates of the UAVs were randomly selected three times on the circle centered at \(p_{\text{target}}\) with a radius of 2.74 km, and the corresponding metrics were calculated by averaging the results of the three trials. The experimental results of the algorithm are presented in Table \ref{tab:tabx}.
Evidently, as the number of UAVs increases, \(\bar{\text{AE}}\) exhibits a distinct decreasing trend, which indicates that the average path following error of all UAVs can be effectively constrained within 10 m even under wind field disturbances. Meanwhile, the path replanning time is consistently limited to less than 3 s when encountering sudden obstacles, which further demonstrates the excellent real-time responsiveness of the proposed algorithm. In terms of the temporal consistency metric (MD), it can be sufficiently controlled within 20 s, which verifies that the algorithm achieves high-precision temporal coordination among UAVs while satisfying all flight constraints.

Notably, the above conclusions are not restricted by the communication topology constraints of the UAV network, which indirectly validates the superior scalability of our algorithm in complex low-altitude environments with wind disturbances and obstacles.

\section{CONCLUSION}
\vspace*{-.5pc}
In this study, we address the problem of robust cooperative path following for multi-UAV systems operating in complex low-altitude terrain with airflow disturbances. To this end, we propose a robust path following controller that realizes finite-time stabilization of path following errors. For sudden obstacles encountered during flight, we design an optimal waypoint replanning algorithm to enable timely and smooth obstacle avoidance. For the temporal index coordination problem among multiple UAVs, we develop a distributed communication protocol that enables effective coordination across the fleet. Extensive experiments are conducted to verify the robustness, efficiency, and coordination performance of the proposed algorithms, demonstrating their promising potential for practical deployment in complex low-altitude mission scenarios.

%


%

\ifCLASSOPTIONcaptionsoff
  \newpage
\fi



%

\bibliographystyle{unsrt}
\bibliography{ref}

%

%
%
%




\end{document}